\documentclass[11pt]{article}
\usepackage{coling2020}
\usepackage{times}
\usepackage{url}
\usepackage{latexsym}
\usepackage{danudefs}
\usepackage{multirow}
\usepackage{microtype}
\usepackage{hyperref}
\usepackage{wrapfig}
\usepackage{graphicx}
\usepackage{subfigure}

\usepackage{todonotes}

\usepackage{microtype}

\colingfinalcopy 

\usepackage[english]{babel}
\usepackage{hyperref}
\addto\captionsenglish{%
}
\addto\extrasenglish{%
}

\usepackage{xr}
\makeatletter
\newcommand*{\addFileDependency}[1]{
  \typeout{(#1)}
  \@addtofilelist{#1}
  \IfFileExists{#1}{}{\typeout{No file #1.}}
}
\makeatother
\newcommand*{\myexternaldocument}[1]{%
    \externaldocument{#1}%
    \addFileDependency{#1.tex}%
    \addFileDependency{#1.aux}%
}
\myexternaldocument{Supplementary}

\usepackage{todonotes}

\makeatletter
\if@todonotes@disabled

\else

\fi
\makeatother

\usepackage{booktabs}
\usepackage{graphics}
\usepackage{adjustbox}

\newtheorem{lemma}[theorem]{Lemma}

\makeatletter
\newcommand{\tblcaption}[1]{\def\@captype{table}\caption{#1}}
\makeatother

\newcommand{\cov}{\mathbf{\Sigma}_{xx}}

\title{Autoencoding Improves Pre-trained Word Embeddings}

\author{
    Masahiro Kaneko\\
    Tokyo Metropolitan University \\
  {\tt kaneko-masahiro@ed.tmu.ac.jp}
    \And
    Danushka Bollegala\Thanks{ Danushka Bollegala holds concurrent appointments as a Professor at University of Liverpool and as an Amazon Scholar. This paper describes work performed at the University of Liverpool and is not associated with Amazon.} \\
  University of Liverpool, Amazon\\
  {\tt danushka@liverpool.ac.uk}}

\date{}

\begin{document}
\maketitle
\begin{abstract}
  Prior work investigating the geometry of pre-trained word embeddings have shown that word embeddings to be distributed in a narrow cone and by 
 centering and projecting using principal component vectors one can increase the accuracy of a given set of pre-trained word embeddings.
 However, theoretically this post-processing step 
 is equivalent to applying a linear autoencoder to minimise the squared $\ell_{2}$ reconstruction error.
 This result contradicts prior work~\cite{mu2018allbutthetop} that proposed to \emph{remove} the top principal components from pre-trained embeddings.
 We experimentally verify our theoretical claims and show that retaining the top principal components is indeed useful for improving pre-trained word embeddings, without requiring access to additional linguistic resources or labeled data. 
\end{abstract}

\section{Introduction}
\label{sec:intro}

Pre-trained word embeddings have been successfully used as features for representing input texts in many NLP tasks~\cite{Dhillon:2015,Mnih:HLBL:NIPS:2008,Collobert:2011,Huang:ACL:2012,Milkov:2013,Pennington:EMNLP:2014}. 
\newcite{mu2018allbutthetop} showed  that the accuracy of pre-trained word embeddings can be further improved in a post-processing step, without requiring additional training data, by removing the mean of the word embeddings (\emph{centering}) computed over the set of words (i.e. vocabulary) and projecting onto the directions defined by the principal component vectors, excluding the top principal components.
They empirically showed that pre-trained word embeddings are distributed in a narrow cone around the mean embedding vector, and centering and projection help to reinstate isotropy in the embedding space. 
This post-processing operation has been repeatedly proposed in different contexts such as with distributional (counting-based) word representations~\cite{sahlgren-etal-2016-gavagai} and sentence embeddings~\cite{Arora:ICLR:2017}.

Independently to the above, autoencoders have been widely used for fine-tuning pre-trained word embeddings such as for removing gender bias~\cite{kaneko-bollegala-2019-gender}, meta-embedding~\cite{Bao:COLING:2018}, cross-lingual word embedding~\cite{Wei:IJCAI:2017} and domain adaptation~\cite{Chen:ICML:2012}, to name a few.
However, it is unclear whether better performance is obtained  simply by applying an autoencoder (a self-supervised task, requiring no labelled data) on pre-trained word embeddings, without performing any task-specific fine-tuning (requires labelled data for the task).

A connection between principal component analysis (PCA) and linear autoencoders was first proved by \newcite{Baldi:1989}, extending the analysis by \newcite{Bourlard:1988}.
We revisit this analysis and theoretically prove that one must \emph{retain} the largest principal components instead of removing them as proposed by \newcite{mu2018allbutthetop} in order to minimise the squared $\ell_{2}$ reconstruction loss.

Next, we experimentally show that by applying a non-linear autoencoder we can post-process a given set of pre-trained word embeddings and obtain more accurate word embeddings than by the method proposed by \newcite{mu2018allbutthetop}.
Although \newcite{mu2018allbutthetop} motivated the removal of largest principal components as a method to improve the isotropy of the word embeddings, our empirical findings show that autoencoding automatically improves isotropy.

\section{Autoencoding as Centering and PCA Projection}

Let us consider a set of $n$-dimensional pre-trained word embeddings, $\{\vec{x}_{i}\}_{i=1}^{N}$ for a vocabulary, $\cV$, consisting of $N$ words.
We post-process these pre-trained word embeddings using an autoencoder consisting of a single $p (< n)$ dimensional hidden layer, an encoder (defined $\mat{W}_{e} \in \R^{n \times p}$ and bias $\vec{b}_{e} \in \R^{p}$) and a decoder (defined by $\mat{W}_{d} \in \R^{p \times n}$ and bias $\vec{b}_{d} \in \R^{n}$). 
Let $\mat{X} \in \R^{n \times N}$ be the embedding matrix.
Using matrices $\mat{B} \in \R^{p \times N}$, $\mat{H} \in \R^{p \times N}$ and $\mat{Y} \in \R^{n \times N}$ respectively denoting the activations, hidden states and reconstructed output embeddings, the autoencoder can be specified as follows.
\noindent
\begin{align*}
 \mat{B} = \mat{W}_{e}\mat{X} + \vec{b}_{e}\vec{u}\T, \qquad \mat{H} &= F(\mat{B}), \qquad \mat{Y} = \mat{W}_{d}\mat{H} + \vec{b}_{d}\vec{u}\T
\end{align*}
Here, $\vec{u} \in \R^{N}$ is a vector consisting of ones and $F$ is an element-wise activation function.
The squared $\ell_{2}$ reconstruction loss, $J$, for the autoencoder is given by \eqref{eq:J}.
\noindent
\begin{align}
 \label{eq:J}
 J(\mat{W}_{e}, \mat{W}_{d}, \vec{b}_{e}, \vec{b}_{d}) = 
 \norm{\mat{W}_{d}F(\mat{W}_{e}\mat{X} + \vec{b}_{e}\vec{u}\T) + \vec{b}_{d}\vec{u}\T}^{2}
\end{align}
The reconstruction loss of the autoencoder is given by Lemma~\ref{lem:cent}, proved in the appendix.
\begin{lemma}
\label{lem:cent}
Let $\mat{X}'$ and $\mat{H}'$ respectively denote the centred embedding and hidden state matrices. 
Then, \eqref{eq:J} can be expressed using  $\mat{X}'$ and $\mat{H}'$ as
$J(\mat{W}_{e}, \mat{W}_{d}, \vec{b}_{e}, \hat{\vec{b}}_{d}) = \norm{\mat{X}' - \mat{W}_{d}\mat{H}'}^{2} $,
where the decoder's optimal bias vector is given by $\hat{\vec{b}}_{d} = \frac{1}{N} \left(\mat{X}-\mat{W}_{d}\mat{H}\right)\vec{u}$.
\end{lemma}

Lemma~\ref{lem:cent} holds even for non-linear autoencoders and claims that the centering happens automatically during the minimisation of the reconstruction error.
Following Lemma~\ref{lem:cent}, we can assume that the embedding matrix, $\mat{X}$, to be already centred and can limit further discussions to this case.
Moreover, after centering the input embeddings, the biases can be \emph{absorbed} into the encoder/decoder matrices by setting an extra dimension that is always equal to 1 in the pre-trained word embeddings.
This has the added benefit of simplifying the notations and proofs.
Under these conditions \autoref{th:PCA} shows an important connection between linear autoencoders and PCA. 
\begin{theorem}
\label{th:PCA}
Assume that $\cov = \mat{X}\mat{X}\T$ is full-rank with $n$ distinct eigenvalues $\lambda_{1} > \ldots > \lambda_{n}$.
Let $\cI = \{i_{1}, \ldots, i_{p}\}$ $(1 \leq i_{1} < \ldots < i_{p} \leq n)$ be any ordered $p$-index set, and $\mat{U}_{\cI} = [ \vec{u}_{i_{1}}, \ldots \vec{u}_{i_{p}} ]$
denote the matrix formed by the orthogonal eigenvectors of $\cov$ associated with the eigenvalues $\lambda_{i_{1}}, \ldots, \lambda_{i_{p}}$.
Then, two full-rank matrices $\mat{W}_{d}$ and $\mat{W}_{e}$ define a critical point of \eqref{eq:J} for a linear autoencoder if and only if there exists an 
ordered $p$-index set $\cI$ and an invertible matrix $\mat{C} \in \R^{p \times p}$ such that 
\noindent
\begin{align}
\mat{W}_{d} &= \mat{U}_{\cI} \mat{C} \label{eq:Wd} \\
\mat{W}_{e} &= \mat{C}\inv \mat{U}_{\cI}\inv \label{eq:We} .
\end{align}
Moreover, the reconstruction error, $J(\mat{W}_{e}, \mat{W}_{d})$ can be expressed as
\noindent
\begin{align}
J(\mat{W}_{e}, \mat{W}_{d}) = \tr(\cov) - \sum_{t \in \cI} \lambda_{t} . \label{eq:eig}
\end{align}
\end{theorem}
\noindent
Proof of \autoref{th:PCA} and approximations for non-linear activations are given in the appendix.
Because $\cov$ is a covariance matrix, it is positive semi-definite. 
Strict positivity corresponds to it being full-rank and is usually satisfied in practice for pre-trained word embeddings, which are dense and use a small $n (\ll
N)$ independent dimensions for representing the semantics of the words.
Moreover, $\mat{W}_{e}$, $\mat{W}_{d}$ are randomly initialised in practice making them full-rank as assumed in \autoref{th:PCA}.

The connection between linear autoencoders and PCA was first proved by \newcite{Baldi:1989}, extending the analysis by \newcite{Bourlard:1988}.
Reconstructing the principal component vectors from an autoencoder has been discussed by~\newcite{Plaut:2018} without any formal proofs.
However, to the best of our knowledge, a theoretical justification for post-processing pre-trained word embeddings by autoencoding has not been provided before.

According to \autoref{th:PCA}, we can minimise \eqref{eq:eig} by selecting the largest eigenvalues as $\lambda_{t}$.
This result contradicts the proposal by \newcite{mu2018allbutthetop} to project the word embeddings away from the largest principal component vectors, which is motivated as a method to improve isotropy in the word embedding space.
They provided experimental evidence to the effect that largest principal component vectors encode word frequency and removal of them is not detrimental to semantic tasks such as semantic similarity measurement and analogy detection.
However, the frequency of a word is an important piece of information for tasks that require differentiating stop words and content words such as in information retrieval.
Actually, \newcite{raunak-etal-2020-dimensional} demonstrated that removing the top principal components does not necessarily lead to performance improvement.
Moreover, contextualised word embeddings such as BERT~\cite{BERT} and Elmo~\cite{Elmo} have shown to be anisotropic despite their superior performance in a wide-range of NLP tasks~\cite{ethayarajh-2019-contextual}.
Therefore, it is not readily obvious whether removing the largest principal components to satisfy isotropy is a universally valid strategy.
On the other hand, our experimental results show that by autoencoding not only we obtain better embeddings than \newcite{mu2018allbutthetop}, but also it improves the isotropy of the pre-trained word embeddings.

\section{Experiments}

\begin{wraptable}[10]{L}{0.5\textwidth}
\centering
\vspace{-0.1cm}
\begin{tabular}{lrr}
\toprule
\bf Parameter & \bf Value \\
\midrule
Optimizer           & Adam      \\
Learning rate       & 0.0002    \\
Dropout rate        & 0.2       \\
Batch size          & 256       \\
Activation function & $\tanh$      \\
\bottomrule
\end{tabular}
\caption{Hyperparameter values of the autoencoder.}
\label{tab:params}
\end{wraptable}

To evaluate the proposed post-processing method, we use the following pre-trained word embeddings: 
\textbf{Word2Vec}\footnote{\url{https://code.google.com/archive/p/word2vec/}} (300-dimensional embeddings for ca. 3M words learnt from the Google News corpus), 
\textbf{GloVe}\footnote{\url{https://github.com/stanfordnlp/GloVe}} (300-dimensional word embeddings for ca. 2.1M words learnt from the Common Crawl), 
and \textbf{fastText}\footnote{\url{https://fasttext.cc/docs/en/english-vectors.html}} (300-dimensional embeddings for ca. 2M words learnt from the Common Crawl).

We use the following benchmarks datasets: for semantic similarity \textbf{WS-353}; \newcite{Agirre:ACL:2009}, \textbf{SIMLEX-999}; \newcite{SimLex},  \textbf{RG-65}; \newcite{RG}, \textbf{MTurk-287}; \newcite{Radinsky:WWW:2011}, \textbf{MTurk-771}; \newcite{Halawi:KDD:2012} and \textbf{MEN}; \newcite{MEN}, for analogy \textbf{Google}, \textbf{MSR}~\cite{Milkov:2013}, and \textbf{SemEval}; \newcite{SemEavl2012:Task2}) and for concept categorisation \textbf{BLESS}; \newcite{BLESS:2011} and \textbf{ESSLI}; \newcite{ESSLLI}) to evaluate word embeddings.

\autoref{tab:params} lists the hyperparameters and their values for the autoencoder-based post-processing method used in the experiments. 
We used the syntactic analogies in the \textbf{MSR}; \newcite{Milkov:2013} dataset for setting the hyperparameters.
We input each set of embeddings separately to an autoencoder with one hidden layer and minimise the squared $\ell_{2}$ error using Adam as the optimiser.
The pre-trained embeddings are then sent through the trained autoencoder and its hidden layer outputs are used as the post-processed word embeddings.
We train an autoencoder (denoted as \textbf{AE}) with a $300$-dimensional hidden layer and a $\tanh$ activation.
Moreover, to study the effect of nonlinearities we train the a linear autoencoder (\textbf{LAE}) without using any nonlinear activation functions in its $300$-dimensional hidden layer.
Due to space limitations, we show results for autoencoders with different hidden layer sizes in the appendix.
We compare the embeddings post-processed using \textbf{ABTT} (stands for \emph{all-but-the-top})~\cite{mu2018allbutthetop}, which removes the top principal components from the pre-trained embeddings.

\begin{table*}[t!]
 \small
 \centering
 \begin{adjustbox}{width=\textwidth,center}
 \begin{tabular}{lcccccccccccc}\toprule
 Embedding & \multicolumn{4}{c}{Word2Vec} & \multicolumn{4}{c}{GloVe} & \multicolumn{4}{c}{fastText} \\ 
 \cmidrule(r){2-5} \cmidrule(r){6-9} \cmidrule(r){10-13}
 Dataset & Original & ABTT & LAE & AE & Original & ABTT & LAE & AE & Original & ABTT & LAE & AE \\ \midrule
 	WS-353 		& \textbf{62.4} & 61.2 & 61.8 & 61.8 & 60.6 & 61.5 & 64.0 & \textbf{65.8} & 65.9 & 67.7 & \textbf{69.0} & \textbf{69.0} \\
	SIMLEX-999  & 44.7 & 45.4 &  \textbf{45.5} & \textbf{45.5} & 39.5 & 41.5 & 40.8 & \textbf{42.2} & 46.2 & 47.4 & \textbf{48.8} & \textbf{48.8} \\
	RG-65 		& 75.4 & 76.0 & 76.2 &\textbf{76.3} & 68.1 & 68.0 & 71.4 & \textbf{72.3} & 78.4 &  \textbf{81.4} & 80.4 & 80.5 \\
	MTurk-287 	&  \textbf{69.0} & 68.9 & \textbf{69.0} & 68.9 & 71.8 & 71.9 & 73.6 & \textbf{74.4} & 73.3 & 73.8 & \textbf{74.7} & \textbf{74.7} \\
	MTurk-771 	& 63.1 & 63.7 &  63.8 & \textbf{63.9} & 62.7 & 63.7 & 66.2 & \textbf{67.7} & 69.6 & 71.8 & 72.3 & \textbf{72.4} \\ 
	MEN 		& 68.1 & 68.3 & 69.2 & \textbf{69.3} & 67.7 & 69.5 & 73.0 & \textbf{74.8} & 71.1 & 75.7 & 75.9 & \textbf{76.0} \\ \midrule
	MSR 		&  \textbf{73.6} & 73.2 & 73.5 & 73.4 & 73.8 & 73.2 & 74.3 & \textbf{74.4} & 87.1 &  \textbf{88.0} & 87.3  & 87.3 \\
	Google 		& 74.0 &  \textbf{74.8} & 74.3 & 74.3 & 76.8 & 76.9 & \textbf{77.2} & 77.1 & 85.3 &  \textbf{88.0} & 86.4  & 86.4 \\ 
	SemEval 	& 20.0 &  19.9 & \textbf{20.4} & 20.3 & 15.4 & 17.2 & 17.2 & \textbf{17.6} & 21.0 &  23.2 & 23.2 & \textbf{23.3} \\ \midrule
	BLESS & 70.5 & \textbf{71.0} & 68.5 & 70.0 & 76.5 & 76.5 & 75.0 & \textbf{79.5} & 75.5 & 79.0 & 79.5 & \textbf{80.5} \\
	ESSLI & 75.5 & 73.7 & 73.8 & \textbf{76.2} & 72.2 & 72.2 & \textbf{73.0} & \textbf{73.0} & 74.7 & 76.2 & 76.1 & \textbf{77.0} \\ 
	\bottomrule
 \end{tabular}
 \end{adjustbox}
  \caption{Results are shown for the original embeddings and their post-processed versions by \textbf{ABTT}, linear autoencoder (\textbf{LAE}) and nonlinear autoencoder (\textbf{AE}) for pre-trained Word2Vec, GloVe and fastText embeddings.}
 \label{tbl:sim}
\end{table*}

\autoref{tbl:sim} compares the performance of the \textbf{Original} embeddings against the embeddings post-processed using \textbf{ABTT}, \textbf{LAE} and \textbf{AE}.
For the semantic similarity task, a high degree of Spearman correlation between human similarity ratings and the cosine similarity scores computed using the word embeddings is considered as better.
From \autoref{tbl:sim} we see that \textbf{AE} improves word embeddings and outperforms \textbf{ABTT} in almost all semantic similarity datasets.
For the word analogy task, we use the PairDiff method~\cite{Levy:CoNLL:2014} to predict the fourth word needed to complete a proportional analogy and the accuracy of the prediction is reported.
For the word analogy task, we see that for the GloVe embeddings \textbf{AE} reports the best performance but \textbf{ABTT} performs better for fastText.
Overall, the improvements due to post-processing are less prominent in the word analogy task.
This behaviour was also observed by \newcite{mu2018allbutthetop} and is explained by the fact that analogy solving is done using vector difference, which is not influenced by centering.

In the concept categorisation task, we measure the Euclidean distance between two words, computed using their embeddings as the distance measure, and use the $k$-means clustering algorithm to group words into clusters separately in each benchmark dataset.
Cluster purity~\cite{Manning:IR} is computed as the evaluation measure using the gold category labels provided in each benchmark dataset.
High values of purity would indicate that the word embeddings capture information related to the semantic classes of words.
From \autoref{tbl:sim} we see that \textbf{AE} outperforms \textbf{ABTT} in all cases, except on BLESS with Word2Vec embeddings.

From \autoref{tbl:sim} we see that for the pre-trained GloVe and fastText embeddings, both linear (\textbf{LAE}) and non-linear autoencoders (\textbf{AE}) yield consistently better post-processed embeddings than the original embeddings.
For the pre-trained Word2Vec embeddings, we see that using \textbf{LAE} or \textbf{AE} produces better embeddings in seven out of the eleven benchmark datasets.
However, according to $p < 0.05$ Fisher transformation we see that the performance difference between \textbf{LAE} and \textbf{AE} is not statistically significant in most datasets.
Considering the theoretical equivalence between PCA and linear autoencoders, this result shows that it is more important to perform centering and apply PCA rather than using a non-linear activation in the hidden layer of the autoencoder.

\begin{wraptable}[]{L}{0.5\textwidth}
\centering
\vspace{-0.1cm}
\begin{tabular}{lcccc}
\toprule
         & Original & ABTT  & LAE   & AE \\
\midrule
Word2Vec & 0.489    & 0.981 & 0.963 & 0.976       \\
GloVe    & 0.018    & 0.943 & 0.782 & 0.884       \\
fastText & 0.773    & 0.995 & 0.992 & 0.990       \\
\bottomrule
\end{tabular}
\tblcaption{The measure of isotropy of original embeddings and after post-processed using \textbf{ABTT} and \textbf{AE}.}
\label{tbl:isotropy}
\end{wraptable}

Following the definition given by \newcite{mu2018allbutthetop}, we empirically estimate the isotropy of a set of embeddings as 
$\gamma = \frac{\min_{\vec{c} \in \cC} Z(\vec{c})}{\max_{\vec{c} \in \cC} Z(\vec{c})}$, where $\cC$ is the set of principal component vectors computed for the given set of pre-trained word embeddings and $Z(\vec{c}) = \sum_{x \in \cV} \exp(\vec{c}\T \vec{x})$ is the normalisation coefficient in the partition function defined by \newcite{Arora:TACL:2016}.
$\gamma$ values close to one indicate a high level of isotropy in the embedding space.
From Table~\ref{tbl:isotropy} we see that compared to the original embeddings \textbf{ABTT}, \textbf{LAE} and \textbf{AE} all improve isotropy.

\begin{figure}[t]
  \begin{center}
    \subfigure[Word2Vec]{\label{fig:word2vec}
        \includegraphics[scale=0.68]{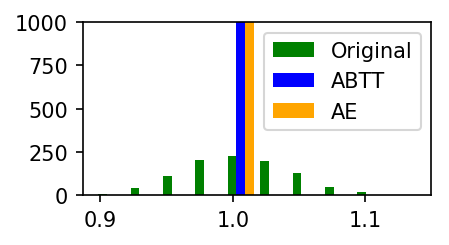}}
    \subfigure[GloVe]{\label{fig:glove}
        \includegraphics[scale=0.68]{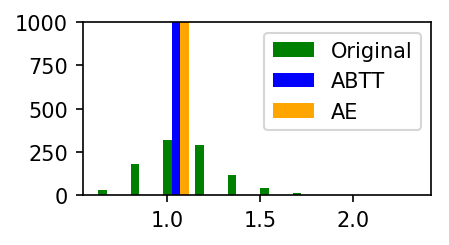}} 
    \subfigure[fastText]{\label{fig:fasttext}
        \includegraphics[scale=0.68]{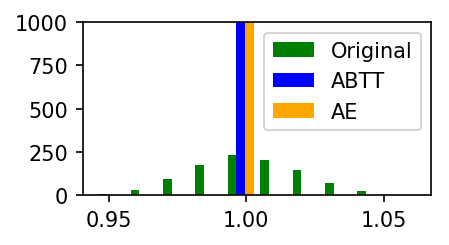}}
  \end{center}
  \caption{The histogram of $Z(c)$ on Word2Vec, GloVe and fastText for 1,000 random vectors $\vec{c}$ of unit norm. The x-axis is normalised by the mean of the values.}
  \label{fig:histogram}
\end{figure}

An alternative approach to verify isotropy is to check whether $Z(c)$ is a constant independent of $c$, which is also known as  the \emph{self-normalisation} property~\cite{andreas-klein:2015:NAACL-HLT}.
\autoref{fig:histogram} shows the histogram of $Z(c)$ of the original pre-trained embeddings, post-processed embeddings using ABTT and AE for pre-trained (a) Word2Vec, (b) GloVe and (c) fastText embeddings for a set of randomly chosen 1000 words $c$ with unit $\ell_2$ norm.
Horizontal axes are normalized by the mean of the values.
From \autoref{fig:histogram}, we see that the original word embeddings in all Word2Vec, GloVe and fastText are far from being isotropic.
On the other hand, AE word embeddings are isotorpic, similar to ABTT word embeddings, in all Word2Vec, GloVe and fastText.
This result shows that isotropy materialises automatically during autoencoding and does not require special processing such as removing the top pricipal components as done by ABTT.

In addition to the theoretical and empirical advantages of autoencoding as a post-processing method, it is also practically attractive.
For example, unlike PCA, which must be computed using the embeddings for all the words in the vocabulary, autoencoders could be run in an online fashion using only a small mini-batch of words at a time.
Moreover, non-linear transformations and regularisation (e.g. in the from of dropout) can be easily incorporated into autoencoders, which can also be stacked for further post-processing.
Although online~\cite{NIPS2006_3144,NIPS2013_5135,Feng:2013} and non-linear~\cite{Scholz_2005} variants of PCA have been proposed, they have not been popular among practitioners due to their computational complexity, scalability and the lack of availability in deep learning frameworks.

\section{Conclusion}
We showed that autoencoding improves pre-trained word embeddings and outperforms the prior proposal for removing top principal components. 
Unlike PCA, which must be computed using the embeddings for all the words in the vocabulary, autoencoders could be run in an online fashion using only a small mini-batch of words at a time.
Moreover, non-linear transformations and regularisation (e.g. in the from of dropout) can be easily incorporated into autoencoders, which can also be stacked for further post-processing.
Although online~\cite{NIPS2006_3144,NIPS2013_5135,Feng:2013} and non-linear~\cite{Scholz_2005} variants of PCA have been proposed, they are less attractive due to computational complexity, scalability and the lack of availability in deep learning frameworks.

\bibliographystyle{coling}
\bibliography{PCA}


\newpage
\appendix
\section{Theoretical Proofs}

The connection between linear autoencoders and principal component analysis (PCA) was first proved by~\cite{Baldi:1989}, which provides the basis for \autoref{th:PCA}. 
The vast applications of autoencoders such as in language~\cite{socher-EtAl:2011:EMNLP,silberer-lapata-2014-learning}, speech~\cite{Gosztolya:2019} and vision~\cite{NIPS2016_6528} domains suggest that non-linear autoencoders can indeed learn better representations than PCA.

In this section, we first show that centering of pre-trained word embeddings happens automatically during the optimisation of an autoencoder.
This result is stated as  Lemma~\ref{lem:cent} in the paper and holds true irrespectively of the activation function used in the autoencoder, including non-linear activation functions.
Next, for linear autoencoders, we recite and prove the connection between linear autoencoders and PCA in the form of \autoref{th:PCA}.
Finally, we discuss the approximations of the \autoref{th:PCA} for non-linear autoencoders.

Recall that we defined the autoencoder as follows:
\begin{align}
 \mat{B} &= \mat{W}_{e}\mat{X} + \vec{b}_{e}\vec{u}\T \label{eq:a_B} \\
 \mat{H} &= F(\mat{B}) \label{eq:a_H} \\
 \mat{Y} &= \mat{W}_{d}\mat{H} + \vec{b}_{d}\vec{u}\T \label{eq:a_Y}
\end{align}
Here, $\vec{u} \in \R^{N}$ is a vector consisting of ones and $F$ is an element-wise activation function.
The squared $\ell_{2}$ reconstruction loss, $J$, for the autoencoder is given by \eqref{eq:a_J}.
\begin{align}
 \label{eq:a_J}
 &J(\mat{W}_{e}, \mat{W}_{d}, \vec{b}_{e}, \vec{b}_{d}) = \nonumber \\
 &\norm{\mat{W}_{d}F(\mat{W}_{e}\mat{X} + \vec{b}_{e}\vec{u}\T) + \vec{b}_{d}\vec{u}\T}^{2} 
\end{align}

For such an autoencoder, Lemma~\ref{lem:cent} holds.

\vspace{0.1cm}
\noindent
{\bf Lemma 1.}
{\it Let $\mat{X}'$ and $\mat{H}'$ respectively denote the centred embedding and hidden state matrices. 
Then, \eqref{eq:a_J} can be expressed using  $\mat{X}'$ and $\mat{H}'$ as
$J(\mat{W}_{e}, \mat{W}_{d}, \vec{b}_{d}, \hat{\vec{b}}_{d}) = \norm{\mat{X}' - \mat{W}_{d}\mat{H}'}^{2} $,
where the optimal decoder bias is $\hat{\vec{b}}_{d} = \frac{1}{N} \left(\mat{X}-\mat{W}_{d}\mat{H}\right)\vec{u}$.}
\begin{proof}
Note that the squared $\ell_{2}$ reconstruction error can be written as in \eqref{eq:a_Jxy}.
\begin{align}
 \label{eq:a_Jxy}
	J(\mat{W}_{e}, \mat{W}_{d}, \vec{b}_{e}, \vec{b}_{d}) &= \norm{\mat{X} - \mat{Y}}^{2}
\end{align}
Substituting for $\mat{Y}$ from \eqref{eq:a_Y} in \eqref{eq:a_Jxy} we have
\begin{align}
 &J(\mat{W}_{e}, \mat{W}_{d}, \vec{b}_{e}, \vec{b}_{d}) = \norm{\mat{X} - \mat{W}_{d}\mat{H} - \vec{b}_{d}\vec{u}\T}^{2}  \label{eq:a_J2}\\
 &= \tr \left( (\mat{X}- \mat{W}_{d}\mat{H} - \vec{b}_{d}\vec{u}\T)\T  (\mat{X}- \mat{W}_{d}\mat{H} - \vec{b}_{d}\vec{u}\T) \right) \label{eq:a_Jtr}
\end{align}

From the definition, $\vec{u} \in \R^{N}$ is a vector with all elements set to 1, where $N$ is the total number of words in the vocabulary $\cV$ for which we are given pre-trained embeddings, arranged as columns in $\mat{X} \in \R^{n \times N}$.
The minimiser of $J$, w.r.t. $\vec{b}_{d}$, $\hat{\vec{b}}_{d}$ satisfies $\frac{\partial J}{\partial \vec{b}_{d}} = \vec{0}$, and is given by \eqref{eq:a_opt-bd}.
\begin{align}
\left(\mat{X} - \mat{W}_{d}\mat{H} - \vec{b}_{d}\vec{u}\T \right) \vec{u} = \vec{0}  \label{eq:a_J:bd-part} \\
\hat{\vec{b}}_{d} = \frac{1}{N} \left( \mat{X} - \mat{W}_{d}\mat{H} \right) \vec{u} \label{eq:a_opt-bd}
\end{align}
In \eqref{eq:a_opt-bd} we used $\vec{u}\T\vec{u} = N$.
Substituting this minimiser $\hat{\vec{b}}_{d}$ back in \eqref{eq:a_J2} we obtain the following.
\begin{align}
 &J(\mat{W}_{e}, \mat{W}_{d}, \vec{b}_{e}, \hat{\vec{b}}_{d}) \nonumber \\
 &= \norm{\mat{X} - \mat{W}_{d}\mat{H} - \frac{1}{N} \left( \mat{X} - \mat{W}_{d}\mat{H} \right) \vec{u}\vec{u}\T}^{2} \\
 &= \norm{\left( \mat{X} - \frac{1}{N}\mat{X}\vec{u}\vec{u}\T \right) - \mat{W}_{d} \left( \mat{H} - \frac{1}{N} \mat{H} \vec{u}\vec{u}\T \right)}^{2} \\
 &= \norm{\left( \mat{X} - \vec{\mu}_{X}\vec{u}\T \right) - \mat{W}_{d} \left( \mat{H} - \vec{\mu}_{H} \vec{u}\T \right)}^{2} \label{eq:a_J-mean}\\
 &= \norm{\mat{X}' - \mat{W}_{d}\mat{H}'}^{2} \label{eq:a_primes}
\end{align}
In \eqref{eq:a_J-mean} we use the mean vectors of embeddings, $\vec{\mu}_{X}$, and hidden states $\vec{\mu}_{H}$ given respectively by \eqref{eq:a_X-mean} and \eqref{eq:a_H-mean}.
\begin{align}
 \vec{\mu}_{X} &= \frac{1}{N} \mat{X}\vec{u} \label{eq:a_X-mean} \\
 \vec{\mu}_{H} &=  \frac{1}{N} \mat{H}\vec{u} \label{eq:a_H-mean}
 \end{align}
 Moreover, we defined the mean-subtracted (i.e. \emph{centred}) versions of $\mat{X}$ and $\mat{H}$ in \eqref{eq:a_primes} respectively by $\mat{X}'$ and $\mat{H}'$ defined as follows.
\begin{align}
 \mat{X}' &= \mat{X} - \vec{\mu}_{X}\vec{u}\T \\
 \mat{H}' &= \mat{H} - \vec{\mu}_{H} \vec{u}\T
\end{align}
\end{proof}

Using Lemma~\ref{lem:cent}, we can replace the embedding matrix, $\mat{X}$, by its pre-centred version and further drop the biases as they can be absorbed into the encoder/decoder weight matrices by introducing a dimension set to 1 in the input and output embeddings.
\autoref{th:PCA}  holds under these transformations.

\vspace{0.1cm}
\noindent
{\bf Theorem 2.}
{\it
Assume that $\cov$ is full-rank with $n$ distinct eigenvalues $\lambda_{1} > \ldots > \lambda_{n}$.
Let $\cI = \{i_{1}, \ldots, i_{p}\}$ $(1 \leq i_{1} < \ldots < i_{p} \leq n)$ is any ordered $p$-index set, and $\mat{U}_{\cI} = [ \vec{u}_{i_{1}}, \ldots \vec{u}_{i_{p}} ]$
denote the matrix formed by the orthogonal eigenvectors of $\cov$ associated with the eigenvalues $\lambda_{i_{1}}, \ldots, \lambda_{i_{p}}$.
Then two full-rank matrices $\mat{W}_{d}$ and $\mat{W}_{e}$ define a critical point of \eqref{eq:a_J} for a linear autoencoder if and only if there exists an 
ordered $p$-index set $\cI$ and an invertible matrix $\mat{C} \in \R^{p \times p}$ such that 
\begin{align}
\mat{W}_{d} &= \mat{U}_{\cI} \mat{C} \label{eq:a_Wd} ,\\
\mat{W}_{e} &= \mat{C}\inv \mat{U}_{\cI}\inv \label{eq:a_We} .
\end{align}
Moreover, the reconstruction error, $J(\mat{W}_{e}, \mat{W}_{d})$ can be expressed as
\begin{align}
J(\mat{W}_{e}, \mat{W}_{d}) = \tr(\cov) - \sum_{t \in \cI} \lambda_{t} . \label{eq:a_eig}
\end{align}
}

\begin{proof}
The squared $\ell_{2}$ reconstruction error can be written by dropping the bias terms and using the centred word embedding matrix $\mat{X}$ as in \eqref{eq:a_Jcent}.
\begin{align}
 &J(\mat{W}_{e}, \mat{W}_{d}) \nonumber \\
 &= \tr \left( \left(\mat{X} - \mat{W}_d \mat{W}_e \mat{X} \right)\T\left(\mat{X} - \mat{W}_d \mat{W}_e \mat{X} \right) \right)  \\
 &=   \tr(\mat{X}\mat{X}\T) - 2\tr(\mat{X}\mat{X}\T\mat{W}_{d}\mat{W}_{e}) \nonumber \\
 &+ \tr(\mat{X}\T\mat{W}_{e}\T\mat{W}_{d}\T\mat{W}_{d}\mat{W}_{e}\mat{X}) \label{eq:a_Jcent} 
\end{align}
When $\mat{W}_{e}$ and $\mat{W}_{d}$ are critical points, by setting $\frac{\partial J}{\partial \mat{W}_{e}} = \mat{0}$ we obtain
\begin{align}
 -2\mat{W}_{d}\T\mat{X}\mat{X}\T + 2 \mat{W}_{d}\T\mat{W}_{d}\mat{W}_{e}\mat{X}\mat{X}\T = \mat{0} \\
(\mat{W}_{d}\T - \mat{W}_{d}\T\mat{W}_{d}\mat{W}_{e})\mat{X}\mat{X}\T = \mat{0} 
\end{align}
Because the covariance matrix for the pre-trained embeddings, $\cov = \mat{X}\mat{X}\T$ is full-rank and is thus invertible, $\mat{W}_{e}$ is given by \eqref{eq:a_We}.
\begin{align}
 \mat{W}_{e} = (\mat{W}_{d}\T\mat{W}_{d})\inv \mat{W}_{d}\T \label{eq:a_We:hat}
\end{align}
Likewise, from  $\frac{\partial J}{\partial \mat{W}_{d}} = \mat{0}$ we obtain
\begin{align}
 \mat{0} &= -2\mat{X}\mat{X}\T\mat{W}_{e}\T + 2\mat{W}_{d}(\mat{W}_{e}\mat{X})(\mat{W}_{e}\mat{X})\T & \\
 \mat{W}_{d} &= \mat{X}\mat{X}\T \mat{W}_{e}\T (\mat{W}_{e} \mat{X} \mat{X}\T \mat{W}_{e}\T) \inv \\
  \mat{W}_{d} &= \cov \mat{W}_{e}\T (\mat{W}_{e} \cov \mat{W}_{e}\T) \inv \label{eq:a_Wd:hat} 
\end{align}

We will first show that \eqref{eq:a_Wd} and \eqref{eq:a_We} satisfy respectively \eqref{eq:a_Wd:hat} and \eqref{eq:a_We:hat}, thereby proving the necessary condition.
Specifically, from \eqref{eq:a_We:hat} and \eqref{eq:a_Wd} we have the following.
\begin{align}
 &(\mat{W}_{d}\T\mat{W}_{d})\inv \mat{W}_{d}\T \nonumber \\
 &= \left( (\mat{U}_{\cI}\mat{C})\T (\mat{U}_{\cI}\mat{C})\right)\inv (\mat{U}_{\cI}\mat{C})\T \\
 &= \left(\mat{C}\T\mat{U}_{\cI}\T\mat{U}_{\cI}\mat{C} \right)\inv \mat{C}\T\mat{U}_{\cI}\T \label{eq:a_U1}\\ 
 &= \mat{C}\inv (\mat{C}\T)\inv \mat{C}\T\mat{U}_{\cI}\inv \\
 &= \mat{C}\inv \mat{U}_{\cI} = \mat{W}_{e}
\end{align}
In \eqref{eq:a_U1}, from the orthogonality of $\mat{U}_{\cI}$, we used $\mat{U}_{\cI}\T\mat{U}_{\cI} = \mat{I}$, where $\mat{I} \in \R^{p \times p}$ is the identity matrix.

Likewise, from \eqref{eq:a_Wd:hat} and \eqref{eq:a_We} we have the following.
\par\nobreak
\begin{align}
 &\cov \mat{W}_{e}\T (\mat{W}_{e}\cov\mat{W}_{e}\T)\inv \nonumber \\
 &= \cov \left(\mat{C}\inv \mat{U}_{\cI}\inv\right)\T \left( \mat{C}\inv \mat{U}_{\cI}\inv \cov (\mat{U}_{\cI}\inv)\T (\mat{C}\inv)\T \right)\inv \\
 &= \cov (\mat{U}_{\cI}\T)\inv (\mat{C}\T)\inv \mat{C}\T \mat{U}_{\cI}\T \cov\inv \mat{U}_{\cI} \mat{C} \\
 &= \mat{U}_{\cI}\mat{C} = \mat{W}_{d}
 \end{align}
 This completes the proof for the necessary condition.
 
 Next, let us look at the sufficient condition.
 For this purpose, we define for a matrix $\mat{M} \in \R^{n \times p}$ $(p \ll n)$ the orthogonal projection onto the subspace spanned by the columns of $\mat{M}$ by $\mat{P}_{\mat{M}}$ given by \eqref{eq:a_PM}.
\begin{align}
 \mat{P}_{\mat{M}} \defeq \mat{M} (\mat{M}\T\mat{M})\inv \mat{M}\T \label{eq:a_PM}
\end{align}
Therefore, for an orthogonal matrix $\mat{U}$, we can evaluate $\mat{P}_{\mat{U}\T\mat{W}_{d}}$ as follows.
\begin{align}
 &\mat{P}_{\mat{U}\T\mat{W}_{d}} \nonumber \\
 &= \mat{U}\T\mat{W}_{d}\left( (\mat{U}\T\mat{W}_{d})\T (\mat{U}\T\mat{W}_{d}) \right)\inv \left(\mat{U}\T\mat{W}_{d}\right)\T \\
 &= \mat{U}\T\mat{W}_{d} \left(\mat{W}_{d}\T \mat{U}\mat{U}\T \mat{W}_{d} \right)\inv \mat{W}_{d}\T\mat{U} \\
 &= \mat{U}\T \mat{W}_{d}\left(\mat{W}_{d}\T\mat{W}_{d}\right)\inv \mat{W}_{d}\T\mat{U} \label{eq:a_s1} \\
 &= \mat{U}\T \mat{P}_{\mat{W}_{d}} \mat{U} \label{eq:a_s2}
\end{align}
In \eqref{eq:a_s2}, from the definition in \eqref{eq:a_PM} we used $\mat{P}_{\mat{W}_{d}} = \mat{W}_{d}\left(\mat{W}_{d}\T\mat{W}_{d}\right)\inv\mat{W}_{d}\T$.
From \eqref{eq:a_s2} we can write $\mat{P}_{\mat{W}_{d}}$ as given in \eqref{eq:a_Pwd}.
\begin{align}
 \label{eq:a_Pwd}
 \mat{P}_{\mat{W}_{d}} = \mat{U} \mat{P}_{\mat{U}\T\mat{W}_{d}} \mat{U}\T
\end{align}

On the other hand, from \eqref{eq:a_Wd:hat} we have
\begin{align}
  &\mat{W}_{d} = \cov \mat{W}_{e}\T (\mat{W}_{e} \cov \mat{W}_{e}\T) \inv ,\\
  &\mat{W}_{d} \left( \mat{W}_{e} \cov \mat{W}_{e}\T \right) = \cov \mat{W}_{e}\T . \label{eq:a_s3}
\end{align}
Right multiplying both sides in \eqref{eq:a_s3} by $\mat{W}_{d}\T$ we arrive at
\begin{align}
 \mat{W}_{d} \mat{W}_{e} \cov \mat{W}_{e}\T \mat{W}_{d}\T &= \cov \left(\mat{W}_{d}\mat{W}_{e}\right)\T . \label{eq:a_s4}
\end{align}
However, by left multiplying \eqref{eq:a_We:hat} by $\mat{W}_{d}$ we get
\begin{align}
 \mat{W}_{d}\mat{W}_{e} = \mat{W}_{d}(\mat{W}_{d}\T\mat{W}_{d})\inv \mat{W}_{d}\T = \mat{P}_{\mat{W}_{d}} . \label{eq:a_prod}
 \end{align}
 Substituting for $\mat{W}_{d}\mat{W}_{e}$ from \eqref{eq:a_prod} back in \eqref{eq:a_s4} we obtain
\begin{align}
 \mat{P}_{\mat{W}_{d}}  \cov \mat{P}_{\mat{W}_{d}}\T = \cov \mat{P}_{\mat{W}_{d}}\T .\label{eq:a_s5}
\end{align}
Note that by the definition in \eqref{eq:a_PM}, $\mat{P}_{\mat{W}_{d}}$ is symmetric (i.e. $\mat{P}_{\mat{W}_{d}}\T = \mat{P}_{\mat{W}_{d}}$).
Therefore, \eqref{eq:a_s5} can be further simplified as given by \eqref{eq:a_s6}.
\begin{align}
 \mat{P}_{\mat{W}_{d}} \cov \mat{P}_{\mat{W}_{d}} = \cov \mat{P}_{\mat{W}_{d}} \label{eq:a_s6}
\end{align}
Taking the transpose of both sides in \eqref{eq:a_s6} we can further show that
\begin{align}
 \left(  \mat{P}_{\mat{W}_{d}} \cov \mat{P}_{\mat{W}_{d}} \right)\T &= \left(  \cov \mat{P}_{\mat{W}_{d}} \right)\T \\
 \mat{P}_{\mat{W}_{d}} \cov \mat{P}_{\mat{W}_{d}}  &= \cov  \label{eq:a_s7}
\end{align}
From \eqref{eq:a_s6} and \eqref{eq:a_s7} we can deduce that
\begin{align}
  \mat{P}_{\mat{W}_{d}} \cov = \cov  \mat{P}_{\mat{W}_{d}} =  \mat{P}_{\mat{W}_{d}} \cov  \mat{P}_{\mat{W}_{d}} \label{eq:a_symm}
\end{align}

Because $\cov$ is real and symmetric, it can be diagonalised using an orthogonal matrix $\mat{U}$, containing the eigenvectors $\vec{u}_{1}, \ldots, \vec{u}_{n}$ of $\cov$ in columns, corresponding to the eigenvalues $\lambda_{1}, \ldots, \lambda_{n}$.
Specifically, we can write this as in \eqref{eq:a_eig}.
\begin{align}
 \cov = \mat{U} \mathbf{\Lambda} \mat{U}\T \label{eq:a_eig}
\end{align}
Here, $\mathbf{\Lambda}$ is a diagonal matrix with non-increasing eigenvalues $\lambda_{1}, \ldots, \lambda_{n}$.

Let us substitute for $\mat{P}_{\mat{W}_{d}}$ from \eqref{eq:a_Pwd} and for $\cov$ from \eqref{eq:a_eig} in \eqref{eq:a_symm}.
\begin{align}
  \mat{P}_{\mat{W}_{d}} \cov &= \cov  \mat{P}_{\mat{W}_{d}} \\
  \mat{U} \mat{P}_{\mat{U}\T\mat{W}_{d}} \mat{U}\T \mat{U} \mathbf{\Lambda} \mat{U}\T &= \mat{U} \mathbf{\Lambda} \mat{U}\T  \mat{U} \mat{P}_{\mat{U}\T\mat{W}_{d}} \mat{U}\T \\
    \mat{U} \mat{P}_{\mat{U}\T\mat{W}_{d}} \mathbf{\Lambda} \mat{U}\T &= \mat{U} \mathbf{\Lambda}  \mat{P}_{\mat{U}\T\mat{W}_{d}} \mat{U}\T \\
      \mat{P}_{\mat{U}\T\mat{W}_{d}} \mathbf{\Lambda} &= \mathbf{\Lambda}  \mat{P}_{\mat{U}\T\mat{W}_{d}} \label{eq:a_d1}
\end{align}
Because $\lambda_{1} > \ldots > \lambda_{n}$,  $\mat{P}_{\mat{U}\T\mat{W}_{d}}$ must be a diagonal matrix to satisfy \eqref{eq:a_d1}.
Specifically, $\mat{P}_{\mat{U}\T\mat{W}_{d}}$ contains $1$ as an eigenvalue ($p$ times) and $0$ ($n-p$ times), and can be written using a diagonal matrix $\mat{I}_{\cI}$ as follows.
\begin{align}
	\label{eq:a_d2}
 \mat{P}_{\mat{U}\T\mat{W}_{d}} = \mat{I}_{\cI}
\end{align}
where the $(i,i)$ diagonal element of $\mat{I}_{\cI}$ is given by \eqref{eq:a_I}.
\begin{align}
\label{eq:a_I}
 \left(\mat{I}_{\cI}\right)_{(i,i)} =  
 	\begin{cases}
		1 & \text{if } i \in \cI \\
		0 & \text{otherwise}
	\end{cases}
\end{align}

Therefore, there exists a unique index set $\cI = \{i_{1}, \ldots, i_{p}\}$ with $(1 \leq i_{1} < \ldots < i_{p} \leq n)$ such that $\mat{P}_{\mat{U}\T\mat{W}_{d}}$ is a diagonal matrix as given by \eqref{eq:a_d2}. 

From  \eqref{eq:a_Pwd} and \eqref{eq:a_prod} we can write,
\begin{align}
 \mat{P}_{\mat{W}_{d}} &= \mat{U}\mat{P}_{\mat{U}\T\mat{W}_{d}}\mat{U}\T \nonumber \\
 &= \mat{U} \mat{I}_{\cI} \mat{U}\T = \mat{U}_{\cI} \mat{U}_{\cI}\T = \mat{W}_{d}\mat{W}_{e}, \label{eq:a_r1}
\end{align}
where $\mat{U}_{\cI} = [ \vec{u}_{i_{1}}, \ldots \vec{u}_{i_{p}} ]$.
Therefore, $\mat{P}_{\mat{W}_{d}}$ is the orthogonal projection onto the subspace spanned by the columns of $\mat{U}_{\cI}$.
Since the column space of $\mat{W}_{d}$ coincides with the column space of $\mat{U}_{\cI}$, there exists an invertible $\mat{C} \in \R^{p \times p}$ matrix such that
\begin{align}
 \mat{W}_{d} &= \mat{U}_{\cI}\mat{C},  \\
 \mat{W}_{e} &= \mat{C}\inv {\mat{U}_{\cI}}\inv .
\end{align}
Therefore, the parameters (i.e. encoder and decoder matrices) of the autoencoder is uniquely determined only upto the scaling matrix $\mat{C}$.

Next, we will consider the reconstruction loss.
First, let us substitute $\cov$ in \eqref{eq:a_Jcent} and rearrange the terms inside the traces as follows.
\begin{align}
  J(\mat{W}_{e}, \mat{W}_{d}) &=  \tr(\cov) -2\tr( \cov \mat{W}_{d}\mat{W}_{e}) \nonumber \\
  &+ \tr(\cov \mat{W}_{e}\T\mat{W}_{d}\T\mat{W}_{d}\mat{W}_{e})  \label{eq:a_J3}
\end{align}
In \eqref{eq:a_J3}, we used $\tr(\mat{A}\mat{B}\mat{C}) = \tr(\mat{C}\mat{A}\mat{B}) = \tr(\mat{B}\mat{C}\mat{A})$ when the product of the three matrices $\mat{A}, \mat{B}, \mat{C}$ are suitably defined.

Substituting for the product $\mat{W}_{d}\mat{W}_{e}$ from \eqref{eq:a_r1} in \eqref{eq:a_J3} we obtain,
\begin{align}
  J(\mat{W}_{e}, \mat{W}_{d}) &=  \tr(\cov) -2\tr( \cov \mat{P}_{\mat{W}_{d}}) \nonumber \\
  &+ \tr( \cov \mat{P}_{\mat{W}_{d}}\T  \mat{P}_{\mat{W}_{d}}) . \label{eq:a_r2}
\end{align}
From the definition in \eqref{eq:a_PM}, we see that $\mat{P}_{\mat{W}_{d}}$ is symmetric (hence,  $\mat{P}_{\mat{W}_{d}}\T = \mat{P}_{\mat{W}_{d}}$)
and moreover that $\mat{P}_{\mat{W}_{d}} \mat{P}_{\mat{W}_{d}} = \mat{P}_{\mat{W}_{d}}$. 
Using this fact in \eqref{eq:a_r2} we can rewrite, 
\begin{align}
  J(\mat{W}_{e}, \mat{W}_{d}) =  \tr(\cov) -\tr( \cov \mat{P}_{\mat{W}_{d}}) . \label{eq:a_r3}
\end{align}
Substituting \eqref{eq:a_eig} and \eqref{eq:a_Pwd} in \eqref{eq:a_r3} we obtain, 
\begin{align}
 J(\mat{W}_{e}, \mat{W}_{d}) &= \tr(\cov)  - \tr(\mat{U}\mathbf{\Lambda}\mat{U}\T \mat{U} \mat{P}_{\mat{U}\T\mat{W}_{d}}\mat{U}\T) \\
 &= \tr(\cov) - \tr(\mat{U}\mathbf{\Lambda}\mat{P}_{\mat{U}\T\mat{W}_{d}}\mat{U}\T) \\
 &= \tr(\cov) - \tr(\mathbf{\Lambda} \mat{P}_{\mat{U}\T\mat{W}_{d}} \mat{U}\T\mat{U}) \\
 &=  \tr(\cov) - \tr(\mathbf{\Lambda} \mat{P}_{\mat{U}\T\mat{W}_{d}}) \\
 &= \tr(\cov) - \tr(\mathbf{\Lambda}\mat{I}_{\cI}) \label{eq:a_r4} \\
 & =  \tr(\cov) - \sum_{t \in \cI} \lambda_{t}
\end{align}
In \eqref{eq:a_r4} above we used \eqref{eq:a_d2}. 
For a given set of word embeddings, $\cov$ is fixed.
Therefore, to minimise $J(\mat{W}_{e}, \mat{W}_{d})$ we must select the largest eigenvalues as $\lambda_{t}$ (and their corresponding eigenvectors).
This completes the proof of \autoref{th:PCA}.
\end{proof}

\subsection{Linear approximations to non-linear activation functions}
\label{sec:a_nonlinear}

The autoencoder considered in \autoref{th:PCA} is linear in the sense that the elementwise activation function is assumed to be $\mat{H} = F(\mat{B}) = \mat{B}$.
However, in practice autoencoders are used with nonlinear activation units such as rectified linear units ReLU; \newcite{Nair:ICML:2010}, hyperbolic tangent ($\tanh$) and sigmoid ($\sigma$) functions~\cite{LeCun_2012}. 
Exact analysis of \autoref{th:PCA} in the general case is is complicated due to the non-linearity of the activation functions.
Therefore, instead, we consider first-order linear approximations for the above-mentioned non-linear activation functions.

ReLU is a piece-wise linear function as given by \eqref{eq:a_ReLU}.
\begin{align}
 \label{eq:a_ReLU}
 F_{relu}(x) = 
\begin{cases}
 x & x > 0 \\
 0 & \text{otherwise}
\end{cases}
\end{align}
Therefore, when ReLU is in its active region, it can be seen as a linear unit.

For $\tanh$ and $\sigma$, when $x$ is small, we use the first-order Taylor expansion to obtain linear approximations as follows.
\begin{align}
 F_{\tanh}(x) &= \frac{1}{1 + \exp(-x)} \approx \frac{1}{2} + \frac{1}{4} x \\
 F_{\sigma}(x) &= \frac{\exp(x) - \exp(-x)}{\exp(x) + \exp(-x)} \approx 0 + x
\end{align}
Therefore, in their linear regions close to zero, both $\tanh$ and $\sigma$ behave like linear functions.

As empirically investigated in the main paper, the performance difference in the word embeddings post-processed using linear vs. non-linear autoencoders is statistically insignificant. 
Considering the theoretical equivalence between PCA and linear autoencoders, this result shows that it is more important to perform centering and apply PCA rather than using a non-linear activation in the hidden layer of the autoencoder.

\begin{table*}[t]
 \small
 \centering
 \begin{tabular}{lccccccccccccccc} \toprule
 Embedding & \multicolumn{4}{c}{Word2Vec} & \multicolumn{4}{c}{GloVe} & \multicolumn{4}{c}{fastText} \\ 
 \cmidrule(r){2-5} \cmidrule(r){6-9} \cmidrule(r){10-13}
    Dataset     & original        & 150d          & 300d          & 600d          & original & 150d & 300d          & 600d          & original & 150d          & 300d           & 600d              \\ \midrule
 	WS-353 		& \textbf{62.4}   & \textbf{62.4} & 61.8          & 61.7          & 60.6     & 47.8 & 65.8          & \textbf{66.9} & 65.9     & 68.2          & \textbf{69.0}  & 68.7              \\
	SIMLEX-999  & 44.7            & 40.8          & 45.5          & \textbf{45.8} & 39.5     & 30.8 & 42.2          & \textbf{43.2} & 46.2     & 43.3          & 48.8           & \textbf{49.0}     \\
	RG-65 		& 75.4            & 75.6          & \textbf{76.3} & 75.9          & 68.1     & 68.7 & \textbf{72.3} & 72.2          & 78.4     & 77.6          & \textbf{80.5}  & 80.4              \\
	MTurk-287 	& 69.0            & \textbf{70.0} & 68.9          & 68.5          & 71.8     & 56.8 & 74.4          & \textbf{74.9} & 73.3     & \textbf{75.2} & 74.7           & 74.8              \\
	MTurk-771 	& 63.1            & 62.9          & \textbf{63.9} & 63.8          & 62.7     & 51.5 & 67.7          & \textbf{68.3} & 69.6     & 70.5          & \textbf{72.4}  & 72.3              \\ 
	MEN 		& 68.1            & 67.2          & \textbf{69.3} & 69.2          & 67.7     & 59.8 & 74.8          & \textbf{75.4} & 71.1     & 74.8          & 76.0           & \textbf{76.1}     \\ \midrule
	MSR 		& \textbf{73.6}   & 61.9          & 73.4          & 73.6          & 73.8     & 60.2 & 74.4          & \textbf{74.6} & 87.1     & 83.4          & \textbf{87.3}  & \textbf{87.3}      \\
	Google 		& 74.0            & 68.1          & 74.3          & \textbf{74.4} & 76.8     & 69.6 & 77.1          & \textbf{77.2} & 85.3     & 83.4          & \textbf{86.4}  & 86.3 \\ 
	SemEval 	& 20.0            & 16.4          & 20.3          & 19.7          & 15.4     & 13.5 & \textbf{17.6} & 17.3          & 21.0     & 21.4          & \textbf{23.3}  & 23.0 \\ \midrule
	BLESS       & \textbf{70.5}   & 65.0          & 70.0          & 68.5          & 76.5     & 74.0 & \textbf{79.5} & 78.5          & 75.5     & 80.0          & \textbf{80.5}  & 80.0  \\
	ESSLI       & 75.5            & \textbf{76.2} & \textbf{76.2} & 74.5          & 72.2     & 56.7 & \textbf{73.0} & \textbf{73.0} & 74.7     & 72.4          & \textbf{77.0}  & 76.9  \\	\bottomrule
 \end{tabular}
 \caption{The autoencoder results using 150, 300 and 600 dimensions for the hidden layer in contrast to original word embeddings.} 
 \label{tbl:a_dim}
\end{table*}

\section{Experimental Settings and Additional Results}

We use semantic similarity
(\textbf{WS-353}; \newcite{Agirre:ACL:2009}, \textbf{SIMLEX-999}; \newcite{SimLex},  \textbf{RG-65}; \newcite{RG}, \textbf{MTurk-287}; \newcite{Radinsky:WWW:2011}, \textbf{MTurk-771}; \newcite{Halawi:KDD:2012} and \textbf{MEN}; \newcite{MEN}), analogy (\textbf{Google}, \textbf{MSR}~\cite{Milkov:2013}, and \textbf{SemEval}; \newcite{SemEavl2012:Task2}) and concept categorisation \textbf{BLESS}; \newcite{BLESS:2011} and \textbf{ESSLI}; \newcite{ESSLLI}) benchmark datasets that were already mentioned in the main article for additional experiments. 
Experiments are conducted using the same 300 dimensional pre-trained embedding learnt using Word2Vec, GloVe and fastText as described in the main body of the paper.

To evaluate the effect of the dimensionality of the hidden layer in the autoencoder on the performance of the post-processed embeddings, in \autoref{tbl:a_dim} we train autoencoders with hidden layer dimensionalities of 150, 300 and 600 and compare the performance against the original (non-post-processed) word embeddings.
For the pre-trained embeddings using Word2Vec and fastText, we see that setting the hidden layer's dimensionality to 300, which is equal to the dimensionality of the input word embeddings, produces better results than with 150 or 600 dimensions in the majority of the datasets.
On the other hand, for pre-trained GloVe embeddings we see that overall the performance increases with the dimensionality of the hidden layer, and the best performance is reported with a 600 dimensional hidden layer in the majority of the datasets.

\end{document}